\documentclass[10pt,twocolumn,letterpaper]{article}

\usepackage{cvpr}              
\usepackage{multirow}
\usepackage{pifont}
\usepackage{comment}
\usepackage{graphicx}
\usepackage{colortbl}
\usepackage{algorithm}
\usepackage{amsmath}
\usepackage{amssymb}
\usepackage{booktabs}
\usepackage{subcaption}
\usepackage{newfloat}
\usepackage{listings}
\usepackage{algpseudocode}

\definecolor{cvprblue}{rgb}{0.21,0.49,0.74}
\usepackage[pagebackref,breaklinks,colorlinks,allcolors=cvprblue]{hyperref}

\def\paperID{12126} 
\def\confName{CVPR}
\def\confYear{2025}

\title{Training-Free Zero-Shot Temporal Action Detection\\ with Vision-Language Models}

\author{Chaolei Han\\
Southeast University\\
School of Cyber Science and Engineering\\
{\tt\small chaoleihan@seu.edu.cn}
\and
Hongsong Wang\\
Southeast University\\
School of Computer Science and Engineering\\
{\tt\small hongsongwang@seu.edu.cn}
\and
Jidong Kuang\\
Southeast University\\
School of Cyber Science and Engineering\\
{\tt\small jidongkuang@seu.edu.cn}
\and
Lei Zhang\\
Nanjing Normal University\\
School of Electrical Engineering and Automation\\
{\tt\small leizhang@njnu.edu.cn}
\and
Jie Gui\\
Southeast University\\
School of Cyber Science and Engineering\\
{\tt\small guijie@seu.edu.cn}
}

\begin{document}

\maketitle
\begin{abstract}
Existing zero-shot temporal action detection (ZSTAD) methods predominantly use fully supervised or unsupervised strategies to recognize unseen activities. However, these training-based methods are prone to domain shifts and require high computational costs, which hinder their practical applicability in real-world scenarios.
In this paper, unlike previous works, we propose a training-\textbf{Free} \textbf{Z}ero-shot temporal \textbf{A}ction \textbf{D}etection (FreeZAD) method, leveraging existing vision-language (ViL) models to directly classify and localize unseen activities within untrimmed videos without any additional fine-tuning or adaptation. 
We mitigate the need for explicit temporal modeling and reliance on pseudo-label quality by designing the LOGarithmic decay weighted Outer-Inner-Contrastive Score (LogOIC) and frequency-based Actionness Calibration.
Furthermore, we introduce a test-time adaptation (TTA) strategy using Prototype-Centric Sampling (PCS) to expand FreeZAD, enabling ViL models to adapt more effectively for ZSTAD.
Extensive experiments on the THUMOS14 and ActivityNet-1.3 datasets demonstrate that our training-free method outperforms state-of-the-art unsupervised methods while requiring only 1/13 of the runtime. When equipped with TTA, the enhanced method further narrows the gap with fully supervised methods.

\end{abstract}    
\section{Introduction}
\label{sec:intro}

With the development of social media and surveillance systems, video understanding has become more important. As a fundamental task in video understanding, temporal action detection (TAD) \cite{yang2023basictad,wang2023two} aims to recognize and localize actions in untrimmed videos. Most existing TAD methods are limited to closed-set scenarios, where they can only detect activities that have been observed before, thereby limiting their applicability in real-world scenarios. Open-set TAD \cite{bao2022opental,chen2023cascade}, which has greater significance due to its ability to handle unseen activities, has received insufficient attention.

\begin{figure}[t]
\centering
\includegraphics[width=1\columnwidth]{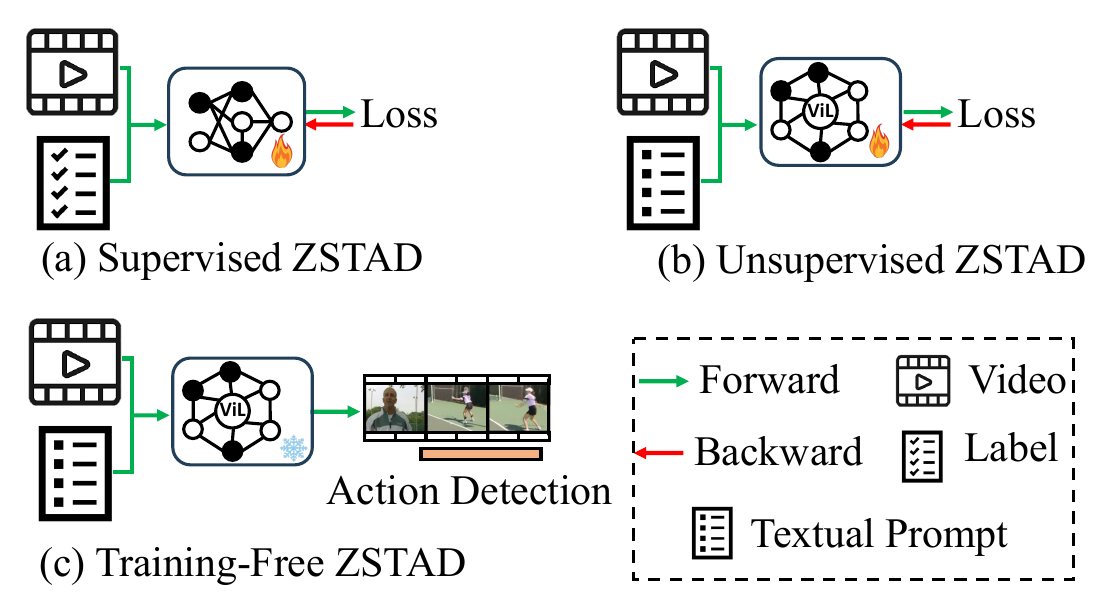} 
\caption{\textbf{We propose the first training-free method for ZSTAD,} distinguishing it from all existing state-of-the-art methods, which are training-based with varying degrees of supervision.}
\label{difference}
\end{figure}

Existing approaches to zero-shot temporal action detection (ZSTAD) can be categorized into two groups based on the type of supervision: fully-supervised methods, which rely on segment-level annotations \cite{zhang2020zstad,zhang2022tn,nag2022zero,li2024detal}, and unsupervised methods, which utilize only unlabeled videos \cite{liberatori2024test}, as shown in Figure \ref{difference}(a) and \ref{difference}(b), respectively.
Whilst more supervision can improve better results, the cost of segment-level annotation is expensive and laborious.
Unsupervised methods, however, require substantial time to learn the data distribution through multiple iterations.

Nearly all existing methods require a training process to detect target actions, thereby introducing several limitations. 
One primary concern is generalization. Models trained on a specific set can hardly address changes in unknown distributions, particularly when faced with domain shifts.
Moreover, action detection data often exhibits class imbalance, with rare or subtle actions being underrepresented in the dataset. This skews the model towards frequent actions, which in turn impacts its generalization capability.
Another important issue is efficiency. Training-based methods typically demand substantial computational resources and storage, especially when handling long-duration or high-frame-rate videos. 
Lastly, data collection for activity-related videos presents challenges in certain domains, particularly those with privacy concerns. Consequently, developing task-specific models for different scenarios is generally impractical.
These considerations motivate us to investigate a novel research question: \textit{Can we develop an approach that is truly training-free to serve the ZSTAD scenario?} 

However, creating such a model is difficult due to two primary factors:
First, unlike traditional methods, training-free approaches cannot use labeled data or learn distributions, leading to a lack of task-specific priors.
Second, the zero-shot setting requires the model to be highly robust, adapting to diverse environments, lighting conditions, camera angles, and activity dynamics.
How to address these two obstacles without training process is key to answering the previous question. 

Owing to their extensive knowledge encapsulation and strong generalization capabilities, ViL models \cite{radford2021learning,jia2021scaling} can effectively address these two problems.
These models acquire strong task-specific priors and generalization capabilities through large-scale multi-task learning, allowing them to quickly adapt to new tasks and make accurate zero-shot predictions without requiring additional explicit training.

To this end, we propose a training-free \textbf{Z}ero-shot temporal \textbf{A}ction \textbf{D}etection (\textbf{FreeZAD}) approach, leveraging existing ViL models to directly classify and localize unseen activities without any additional fine-tuning or adaptation.
Specifically, FreeZAD first generates a video-level pseudo-label by aggregating semantic information extracted from the ViL model, which is then treated as a prototype to guide the action localization.
We mitigate the reliance on explicit temporal modeling by introducing a \textit{LOGarithmic decay weighted Outer-Inner-Contrastive score (LogOIC)}, which is capable of enhancing sensitivity to temporal action boundaries.
To reduce dependence on the quality of pseudo-labels, we propose \textit{Actionness Calibration}, a method that leverages frequency energy from visual features to refine confidence scores, thereby providing a more stable foundation for evaluation. Moreover, we extend the proposed FreeZAD approach by incorporating Test-Time Adaptation (TTA) \cite{liang2024comprehensive} to enable more precise localization by adapting to a single video sequence in an unsupervised manner.

Our contributions can be summarized as follows:
\begin{itemize}
\item To the best of our knowledge, we are the first to investigate the problem of training-free ZSTAD.
\item We propose FreeZAD, a training-free approach for ZSTAD, which effectively leverages the generalization capabilities of ViL models to detect unseen activities.
\item We introduce a simple yet effective TTA method that extends FreeZAD and enhances its performance by enabling adaptation to a video sequence without supervision.
\item Extensive experiments demonstrate that our methods outperform state-of-the-art unsupervised methods and significantly narrow the gap with fully supervised methods.
\end{itemize}

\begin{figure*}[ht]
\centering
\includegraphics[width=2.0\columnwidth]{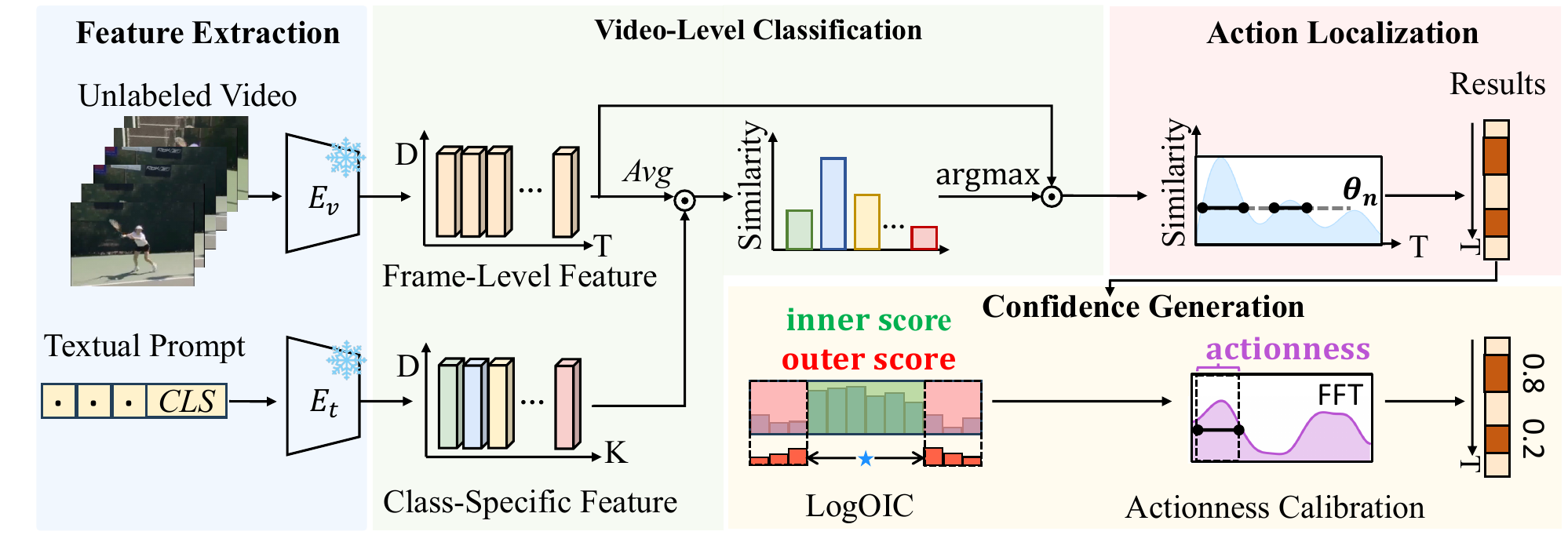} 
\caption{\textbf{Overall architecture of our proposed training-free zero-shot temporal action detection network.} 
The model recognizes and localizes unseen activities within untrimmed videos with only a single forward pass.
Specifically, video-level label is first generated through visual embedding and textual encoding derived from the visual and textual backbones of ViL models.
Then, segments are selected by calculating the similarities between this label and the visual features of each frame through a filtering operation.
Finally, the confidence score for each segment is generated using LogOIC and refined by Actionness Calibration.}
\label{pipeline}
\end{figure*}
\section{Related Work}
\label{sec:relatedwork}

\noindent\textbf{Temporal Action Detection:} Temporal action detection (TAD), also referred to as temporal action localization, requires classification and localization of action instances in untrimmed videos. The fact that the same action may occur multiple times and various actions can happen simultaneously makes the TAD task challenging. Existing TAD work can be categorized into three groups according to the pipeline: 
(1) Two-stage methods \cite{qing2021temporal,tan2021relaxed,xia2022learning,zhao2023movement,zhao2021video} first generate class-agnostic action segments, followed by classifying each candidate segment.
(2) One-stage methods \cite{lin2017single,zhang2022actionformer,shi2023tridet,shao2023action,chen2024video,liu2024end} perform both subtasks simultaneously. 
(3) Query-based methods \cite{kim2024te,zhu2024dual,kim2023self} interact a set of learnable queries with the visual features of videos and directly generate temporal proposals through set matching. These methods are developed in closed-set scenarios. 

To cater for more complicated situations in the real world, many researchers are dedicated to applying the TAD task in open-set scenarios, where action categories are disjoint for training and testing. 
Existing ZSTAD approaches can be further categorized into supervised and unsupervised methods.
For fully-supervised methods, Zhang et al. \cite{zhang2020zstad} encodes seen and unseen activities by taking advantage of Word2Vec \cite{mikolov2013efficient} and successfully captures common semantic information from them. 
TN-ZSTAD \cite{zhang2022tn} enhances label embedding by using a text encoder from CLIP \cite{radford2021learning}, which provides enriched semantic information learned from annotations of abundant image-text pairs. 
Eff-Prompt \cite{ju2022prompting} classifies well-generated category-agnostic proposals by CLIP, demonstrating strong performance in both open-set and closed-set scenarios. 
These two-stage approaches are hampered by mutual disturbances due to the localization error propagation challenge. To address this issue, STALE \cite{nag2022zero} develops a proposal-free framework that leverages the language priors of CLIP, while DeTAL \cite{li2024detal} builds a two-stage network that decouples action localization from classification. 
However, all aforementioned methods primarily rely on laborious and expensive segment-level annotations.
To address this limitation, T3AL \cite{liberatori2024test} attempts an adaptation strategy to detect activity instances without supervision.
All existing work is based on a training process. In contrast, we propose a training-free method that leverages the generalization capabilities of ViL models to classify and localize unseen activities.

\noindent\textbf{ViL models for video understanding:}
The vision-language (ViL) model \cite{alayrac2022flamingo,jia2021scaling,radford2021learning,wang2022ofa,yu2022coca,zhang2024vision} is a multimodal model that aligns and integrates large-scale visual and language data, enabling meaningful associations between visual content and natural language. 
ViL models possess powerful knowledge representation capabilities and have been proven effective for various image tasks, such as detection \cite{gunjal2024detecting,wang2023image}, segmentation \cite{rao2022denseclip,zhou2022extract}, generation \cite{crowson2022vqgan,wang2022clip}, and human-object interaction \cite{wang2024bilateral,zhou2024learning}.

Additionally, some research focuses on transferring knowledge from ViL models to video understanding.
For action recognition, Vita-CLIP \cite{wasim2023vita} proposes a multimodal prompt learning scheme to balance supervised and zero-shot video classification performance within a unified training framework. For video retrieval, STAN \cite{pei2023clipping} extends the ViL model by introducing a branch structure with decomposed spatial-temporal modules, enabling effective temporal modeling for video tasks.
For temporal action localization, Ju et al. \cite{ju2023distilling} propose a novel distillation-collaboration framework that leverages the strengths of a ViL branch to enhance localization accuracy.
All the aforementioned work demonstrates the broad potential of applying ViL models to the video domain.
Inspired by these approaches, we leverage the knowledge encapsulation and generalization capabilities of ViL models to produce comprehensive semantic information from both video and textual prompts.

\noindent\textbf{Test-Time Adaptation:} Test-time adaptation (TTA) is a branch of the domain adaptation (DA) field that aims to adapt a model pre-trained on the source domain to the target domain without requiring access to labels from the source domain \cite{liang2024comprehensive}, making it a safer and more practical approach.
Substantial progress has been made in TTA for various computer vision tasks. In image classification, T3A \cite{iwasawa2021test} adjusts the linear classifier at test time to improve decision boundaries and reduce prediction ambiguity. DomainAdaptor \cite{zhang2023domainadaptor} addresses the domain shift problem by adaptively incorporating statistics in the normalization layer and minimizing a generalized entropy. In video object segmentation, DATTT \cite{liu2024depth} enforces the modulation layer of a single network to predict consistent depth during testing, resulting in significant superiority. ViTTA \cite{Lin_2023_CVPR} demonstrates distribution shifts in video action recognition and addresses it by aligning feature distribution statistics of test set towards training set. 
To the best of our knowledge, T3AL \cite{liberatori2024test} is the only research in the ZSTAD field that directly adapts CoCa \cite{yu2022coca} to a single test video in test time.
While T3AL incorporates additional caption models for post-processing to filter candidate segments, their performance remains limited due to the lack of labeled data guidance and explicit temporal modeling.
Inspired by prior TTA approaches, we redesign an adaptation strategy specifically tailored for image-text alignment, enabling precise localization of activities.

\section{FreeZAD}
\label{sec:freezad}
To flexibly address ZSTAD, we exploit the rich semantic information encoded in ViL models. As illustrated in Figure \ref{pipeline}, we present a training-free approach FreeZAD that requires only a single forward pass of the network. This model consists of three main components: video-level classification, action localization, and confidence generation.

\subsection{Video-Level Classification}
Since no annotation is accessible, a video-level pseudo-label needs to be first generated to serve as a subsequent supervision signal. Let $E_v$ and $E_t$ denote the visual encoder and text encoder of a pre-trained ViL model $\mathcal{M}$, respectively. 
For visual embedding, RGB features $\mathbf{X} \in \mathbb{R}^{T\times D} $ are extracted by $E_v$ from the video at the frame level, where $T$ and $D$ are number of frames and feature dimension, respectively. 
To obtain textual embedding $\mathbf{F}\in \mathbb{R}^{C\times D}$, each class of interest is prompted with the template ``a video of action \textit{CLS}" and then encoded by $E_t$, where $C$ is the number of action categories.

Then, the pseudo-label is generated using the prior knowledge from $\mathcal{M}$, which has been demonstrated to possess strong zero-shot capability. Specifically, the video-level visual feature $\bar{\mathbf{x}}$ is first obtained by averaging $\mathbf{X}$ along the temporal dimension. Thus, $\bar{\mathbf{x}}$ is computed as:
\begin{equation}
    \bar{\mathbf{x}} = \frac{1}{T}\sum_{t=1}^{T}\mathbf{x}_{t} \in \mathbb{R}^{1\times D},
\end{equation}
followed by computing the cosine similarity with the textual encoding. Next, the action category with the highest similarity is selected as the pseudo-label $\hat{c}$. This process can be formulated as follows:
\begin{equation}
    \hat{c} = \mathop{\mathrm{argmax}}\limits_{c\in \mathcal{C}}\frac{\mathbf{f}_{c}\cdot\bar{\mathbf{x}}^T}{\left \| \mathbf{f}_{c} \right \|\left \| \bar{\mathbf{x}} \right \| } ,
\end{equation}
where $\cdot$ denotes the dot production, and $\left \| \cdot  \right \| $ denotes the $L_2$ norm. We view the textual feature of class $\hat{c}$ as prototype $\mathbf{y}$, which is used to guide the subsequent localization task to provide more accurate temporal predictions.

\subsection{Action Localization} 
One of the fundamental requirements of TAD is to identify when the action of interest occurs. To achieve this, $\mathbf{y}$ is used to guide for roughly locating the segments in video where $\hat{c}$ happens. Intuitively, the visual feature of each frame associated with action $\hat{c}$ should have high semantic similarity with its textual feature. Building on this consideration, the cosine similarity is directly computed between $\mathbf{x}_{t}$ and $\mathbf{y}$ as:
\begin{equation}
    s_t = \frac{\mathbf{x}_{t} \cdot \mathbf{y}^T}{\left \| \mathbf{x}_{t} \right \| \left \| \mathbf{y} \right \| }.
\label{eq6}
\end{equation}
To enhance temporal consistency, a moving average of $\mathbf{s}$ should be computed before normalization. 

Once scaled between 0 and 1, the value of $\mathbf{s}_{t}$ can to some extent reflect the probability of feature belonging to activity of interest at the $t$-th frame. Typically, segments can be generated by merging $t$ where $s_{t} > \alpha$, where threshold $\alpha$ is a hyperparameter. 

\subsection{Confidence Generation}

Whilst video-level classification and action localization enable ZSTAD without training, performance remains limited due to the lack of explicit temporal modeling and annotated information. To address these challenges, we propose \textit{LOGarithmic decay weighted Outer-Inner-Contrastive score (LogOIC)} and \textit{Actionness Calibration}, respectively.

\begin{figure}[t]
\centering
\includegraphics[width=1\columnwidth]{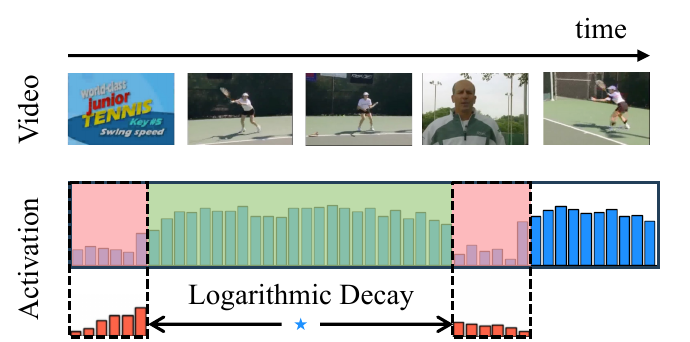} 
\caption{\textbf{An illustration of Logarithmic Decay Weighted Outer-Inner-Contrastive Score (LogOIC).} The green mask and the red mask respectively cover the inner and outer activations within a segment. The adjusted outer activation, shown in orange, are obtained by reweighting the blue activation with a logarithmic decay weight.}
\label{logoic}
\end{figure}

\noindent\textbf{Logarithmic Decay Weighted Outer-Inner-Contrastive Score:} Initially, candidate segments are generated for each pseudo-label $\hat{c}$ by merging frames with similarity scores higher than $\alpha$, with segment-level similarity scores directly used as the confidence scores. However, this metric focuses more on discriminative features while overlooking the quality of action localization, which may result in unstable evaluation. 
The Outer-Inner-Contrastive (OIC) score offers more stable boundary evaluations by constraining the contrast between the inner and outer activation values of candidate segments \cite{shou2018autoloc}. 
Considering the characteristics of temporal information, we hypothesize that the influence of outer activation values on boundary reliability diminishes with increasing distance from the action center. Motivated by these observations, we propose LogOIC, which adjusts weights using a logarithmic decay function for outer activation values to enhance sensitivity to action boundaries, as illustrated in Figure \ref{logoic}. 
The terms $w_{t}^{left}$ and $w_{t}^{right}$ represent the weights of the outer areas preceding and following the candidate segment, which can be calculated as follows:
\begin{align}
        w_{t}^{left} = \frac{1/\log(b-t+\eta )}{ {\textstyle \sum_{m=1}^{l}\log(m+\eta)} }, \\
        w_{t}^{right} = \frac{1/\log(t-e+\eta )}{ {\textstyle \sum_{m=1}^{l}\log(m+\eta)} }, 
\end{align}
where $\eta=1$ is used to control the translation of the logarithmic function and $l=(e-b+1)/4$ is inflation length.

The final LogOIC is formulated as:
\begin{equation}
    p = s_{inner} - \gamma_{1} s_{outer} + \gamma_{2} s_{max},
\end{equation}
where $\gamma_{1}$ and $\gamma_{2}$ are hyperparameters, and $s_{max}$ is the maximum activation level within the segment. The terms $s_{inner}$ and $s_{outer}$ represent the average activation levels within and outside the generated segments, respectively, which can be defined as follows:
\begin{align}
s_{inner} &= \frac{1}{e-b+1} \sum_{t=b}^{e}s_t ,\\
s_{outer} &=\sum_{t=b-l}^{b-1}w^{left}_t s_t + \sum_{t=e+1}^{e+l}w^{right}_t s_t.
\label{equ}
\end{align}



\noindent\textbf{Actionness Calibration:} Generating reliable confidence scores for ranking candidate segments is a challenging task in the field of detection. The previous confidence scores are calculated based on the similarity between visual embedding and textual encoding, making the quality of the segments entirely dependent on the generated prototype, which increases the risk of unreliable rankings.
Frequency, as an inherent property of movement, can naturally reflect movement patterns and regularities to some extent.
While frequency information has been successfully used to enhance accuracy in image classification or action recognition \cite{xu2020learning,xu2023channel}, its application in temporal action detection has been largely underexplored. 
Motivated by their work, we introduce a calibration module for video detection that leverages the frequency information of each frame within a segment to predict the frequency energy, \ie, the actionness score, which is then used to refine the original confidence score.
Specifically, we apply a fast Fourier transform \cite{cooley1965algorithm} to each candidate segment along the temporal dimension. This process is described as:
\begin{equation}
    \mathbf{Z} = \mathcal{F}(\mathbf{X}_{b:e}) \in \mathbb{R}^{(e-b+1) \times D},
\end{equation}
where $\mathcal{F}$ represents fast Fourier transform. Let $z_{t,d}$ be the  element in the $t$-th row and the $d$-th column of $\mathbf{Z}$. The segment-level calibration score is produced by averaging the sum of squares to generate frequency energy. This formula is computed as:
\begin{equation}
    s = \mathit{sigmoid}(\frac{1}{e-b+1}\sum_{t=1}^{e - b + 1}\sum_{d=1}^{D} z_{t,d}^2 ),
\end{equation}
The final confidence score can be refined as follows:
\begin{equation}
        p = p \cdot s.
\end{equation}

Through the elaborately designed modules mentioned above, our model can effectively leverage the ViL models for the ZSTAD task, achieving satisfactory performance even without a training process.

\section{AdaZAD}
\label{sec:adazad}

\begin{figure}[t]
\centering
\includegraphics[width=1\columnwidth]{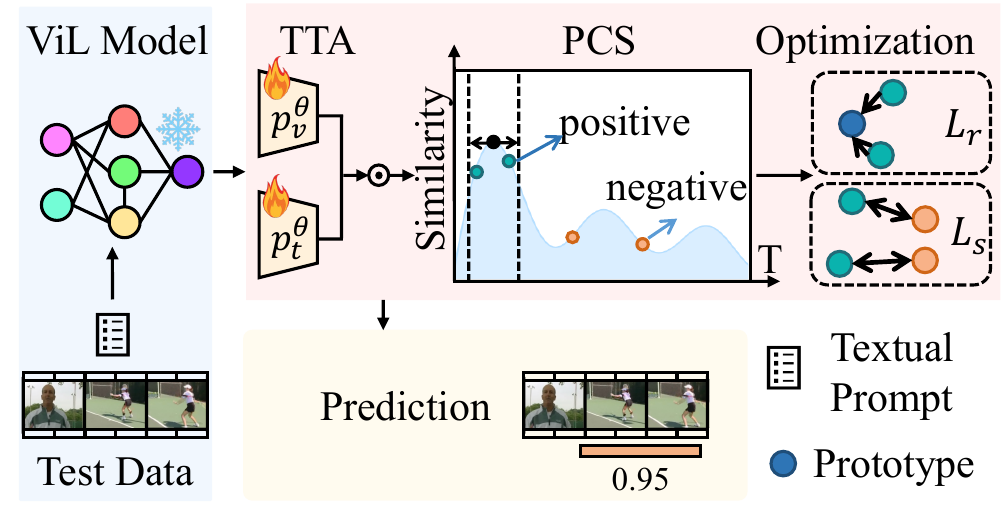} 
\caption{\textbf{Illustration of zero-shot temporal action detection with test-time adaptation (AdaZAD).} This paradigm seeks to adapt the pre-trained ViL models for TAD without annotations, and the detected results are obtained after completing all adaptation steps. Prototype-Centric Sampling (PCS) focuses on selecting positive samples surrounding the highest similarity values to the textual features.}
\label{tta}
\end{figure}
To test the adaptability and extensibility of our FreeZAD, we draw inspiration from previous test-time adaptation (TTA) research \cite{liberatori2024test} and develop an adaptation strategy for ZSTAD, as shown in Figure \ref{tta}. 
In this section, we first outline the general steps of TTA, followed by introducing our \textit{Prototype-Centric Sampling (PCS)}, which is tailored to align visual and textual embeddings more effectively in a common semantic space.

\subsection{General Steps for TTA}
In the top of $E_v$ and $E_t$, $P_v$ and $P_t$ are the projection layers with learnable parameters. 
Once similarity scores of each frame are calculated by Eq.(\ref{eq6}) , \( K \) frames are randomly selected to construct the positive sample visual representations $\mathbf{X}^+ \in \mathbb{R}^{K\times D}$, the positive sample similarities $\mathbf{s}^+ \in \mathbb{R}^{K\times 1}$, and the negative sample similarities $\mathbf{s}^- \in \mathbb{R}^{K\times 1}$, respectively.

The constructed sets are then fed into the contrastive loss function \cite{liberatori2024test} to refine the initial predictions in a self-supervised manner:
\begin{equation}
    \theta^*= \mathop{\mathrm{argmax}}\limits_{\theta}(\beta\mathcal{L}_{r}(\mathbf{X}^+, \mathbf{y})+\mathcal{L}_{s}(\mathbf{s}^+,\mathbf{s}^-,\mathbf{s}_{bin})).
\end{equation}
The learnable parameters \( \theta \) are adapted by simultaneously minimizing representation loss $\mathcal{L}_{r}$ and separation loss $\mathcal{L}_{s}$.
The former loss term is applied to make positive samples closer to the prototype, while the last term keeps the positives and negatives away from each other. In the above formula, $\mathbf{s}_{bin}$ is the idealized similarity vector, with elements being either 0 or 1, and hyperparameter $\beta$ is the loss trade-off coefficient.

\begin{table*}[ht]
\setlength{\tabcolsep}{2.5mm}
\centering
\begin{tabular}{c|l|c|cccccc|ccccc}
\toprule[1pt]
\midrule
\multirow{2}{*}{Data Split} & \multirow{2}{*}{Methods} & \multirow{2}{*}{Label}  & \multicolumn{6}{c|}{THUMOS14}         & \multicolumn{4}{c}{ActivityNet-1.3} \\
                                                 & &    & 0.3    & 0.4   & 0.5   & 0.6   & 0.7   & mAP    & 0.5   & 0.75  & 0.95  & mAP  \\ \midrule
\multirow{7}{*}{\begin{tabular}[c]{@{}c@{}}75\% Seen\\ \\ 25\% Unseen\end{tabular}} 
& Eff-Prompt\cite{ju2022prompting}&\ding{51}                                     & 39.7   & 31.6  & 23.0  & 14.9  & 7.5   & 23.3  & 37.6  & 22.9  & 3.8   & 23.1  \\
& STALE\cite{nag2022zero}&\ding{51}                                              & 40.5   & 32.3  & 23.5  & 15.3  & 7.6   & 23.8  & 38.2  & 25.2  & 6.0   & 24.9  \\
& DeTAL\cite{li2024detal}&\ding{51}                                              & 39.8   & 33.6  & 25.9  & 17.4  & 9.9   & 25.3  & 39.3  & 26.4  & 5.0   & 25.8  \\
& \cellcolor[HTML]{D9D9D9}T3AL\cite{liberatori2024test}&\cellcolor[HTML]{D9D9D9}\ding{55}                  & \cellcolor[HTML]{D9D9D9}19.2   & \cellcolor[HTML]{D9D9D9}12.7  & \cellcolor[HTML]{D9D9D9}7.4   & \cellcolor[HTML]{D9D9D9}4.4   & \cellcolor[HTML]{D9D9D9}2.2   & \cellcolor[HTML]{D9D9D9}9.2       
                                                                                & \cellcolor[HTML]{D9D9D9}28.1   & \cellcolor[HTML]{D9D9D9}14.9  & \cellcolor[HTML]{D9D9D9}3.3   & \cellcolor[HTML]{D9D9D9}15.4  \\
& \cellcolor[HTML]{D9D9D9}Baseline&\cellcolor[HTML]{D9D9D9}\ding{55}                                       & \cellcolor[HTML]{D9D9D9}15.7   & \cellcolor[HTML]{D9D9D9}9.8   & \cellcolor[HTML]{D9D9D9}5.8   & \cellcolor[HTML]{D9D9D9}3.2   & \cellcolor[HTML]{D9D9D9}1.6   & \cellcolor[HTML]{D9D9D9}7.2    
                                                                                & \cellcolor[HTML]{D9D9D9}28.2   & \cellcolor[HTML]{D9D9D9}15.0  & \cellcolor[HTML]{D9D9D9}3.4   & \cellcolor[HTML]{D9D9D9}15.5  \\
& \cellcolor[HTML]{D9D9D9}FreeZAD&\cellcolor[HTML]{D9D9D9}\ding{55}                                            & \cellcolor[HTML]{D9D9D9}21.2   & \cellcolor[HTML]{D9D9D9}13.6  & \cellcolor[HTML]{D9D9D9}8.3   & \cellcolor[HTML]{D9D9D9}4.7   & \cellcolor[HTML]{D9D9D9}2.5   & \cellcolor[HTML]{D9D9D9}10.0
                                                                                & \cellcolor[HTML]{D9D9D9}33.5   & \cellcolor[HTML]{D9D9D9}17.5  & \cellcolor[HTML]{D9D9D9}3.9   & \cellcolor[HTML]{D9D9D9}18.3   \\
& \cellcolor[HTML]{D9D9D9}AdaZAD&\cellcolor[HTML]{D9D9D9}\ding{55}                                         & \cellcolor[HTML]{D9D9D9}\textbf{27.5} & \cellcolor[HTML]{D9D9D9}\textbf{18.7} & \cellcolor[HTML]{D9D9D9}\textbf{12.3} & \cellcolor[HTML]{D9D9D9}\textbf{7.3} & \cellcolor[HTML]{D9D9D9}\textbf{4.0} 
                                                                                & \cellcolor[HTML]{D9D9D9}\textbf{14.0} & \cellcolor[HTML]{D9D9D9}\textbf{34.8} & \cellcolor[HTML]{D9D9D9}\textbf{19.1} & \cellcolor[HTML]{D9D9D9}\textbf{4.1} & \cellcolor[HTML]{D9D9D9}\textbf{19.3}          \\ \midrule
\multirow{7}{*}{\begin{tabular}[c]{@{}c@{}}50\% Seen\\ \\ 50\% Unseen\end{tabular}} & Eff-Prompt\cite{ju2022prompting}&\ding{51}
                                                                                & 37.2   & 29.6  & 21.6  & 14.0  & 7.2   & 21.9  & 32.0  & 19.3  & 2.9   & 19.6  \\
& STALE\cite{nag2022zero}&\ding{51}                                              & 38.3   & 30.7  & 21.2  & 13.8  & 7.0   & 22.2  & 32.1  & 20.7  & 5.9   & 20.5  \\
& DeTAL\cite{li2024detal}&\ding{51}                                              & 38.3   & 32.3  & 24.4  & 16.3  & 9.0   & 24.1  & 34.4  & 23.0  & 4.0   & 22.4  \\
& \cellcolor[HTML]{D9D9D9}T3AL\cite{liberatori2024test}&\cellcolor[HTML]{D9D9D9}\ding{55}                  & \cellcolor[HTML]{D9D9D9}20.7   & \cellcolor[HTML]{D9D9D9}14.3  & \cellcolor[HTML]{D9D9D9}8.9   & \cellcolor[HTML]{D9D9D9}5.3   & \cellcolor[HTML]{D9D9D9}2.7   & \cellcolor[HTML]{D9D9D9}10.4  
                                                                                & \cellcolor[HTML]{D9D9D9}25.8   & \cellcolor[HTML]{D9D9D9}13.9  & \cellcolor[HTML]{D9D9D9}3.1   & \cellcolor[HTML]{D9D9D9}14.3  \\
& \cellcolor[HTML]{D9D9D9}Baseline&\cellcolor[HTML]{D9D9D9}\ding{55}                                       & \cellcolor[HTML]{D9D9D9}15.5   & \cellcolor[HTML]{D9D9D9}9.9   & \cellcolor[HTML]{D9D9D9}5.9   & \cellcolor[HTML]{D9D9D9}3.3   & \cellcolor[HTML]{D9D9D9}1.6   & \cellcolor[HTML]{D9D9D9}7.2   
                                                                                & \cellcolor[HTML]{D9D9D9}26.0   & \cellcolor[HTML]{D9D9D9}14.0  & \cellcolor[HTML]{D9D9D9}3.2   & \cellcolor[HTML]{D9D9D9}14.4  \\
& \cellcolor[HTML]{D9D9D9}FreeZAD&\cellcolor[HTML]{D9D9D9}\ding{55}                                            & \cellcolor[HTML]{D9D9D9}20.7   & \cellcolor[HTML]{D9D9D9}13.4  & \cellcolor[HTML]{D9D9D9}8.3   & \cellcolor[HTML]{D9D9D9}4.7   & \cellcolor[HTML]{D9D9D9}2.5   & \cellcolor[HTML]{D9D9D9}9.9   
                                                                                & \cellcolor[HTML]{D9D9D9}34.1   & \cellcolor[HTML]{D9D9D9}17.9   & \cellcolor[HTML]{D9D9D9}4.0    & \cellcolor[HTML]{D9D9D9}18.7  \\
& \cellcolor[HTML]{D9D9D9}AdaZAD&\cellcolor[HTML]{D9D9D9}\ding{55}                                         & \cellcolor[HTML]{D9D9D9}\textbf{27.6} & \cellcolor[HTML]{D9D9D9}\textbf{19}   & \cellcolor[HTML]{D9D9D9}\textbf{12.4} & \cellcolor[HTML]{D9D9D9}\textbf{7.3} & \cellcolor[HTML]{D9D9D9}\textbf{3.9} 
                                                                                & \cellcolor[HTML]{D9D9D9}\textbf{14.1} & \cellcolor[HTML]{D9D9D9}\textbf{35.7} & \cellcolor[HTML]{D9D9D9}\textbf{20.0} & \cellcolor[HTML]{D9D9D9}\textbf{4.2} & \cellcolor[HTML]{D9D9D9}\textbf{20.0}
                                                                                \\ \midrule
\bottomrule[1pt]
\end{tabular}
\caption{\textbf{Zero-shot TAD performance on THUMOS14 and ActivityNet-1.3 at different tIoU thresholds}, where the best performance without annotations is highlighted in bold. For reference, we also include the methods of standard training and testing with fully annotated data via supervised learning.}
\label{table1}
\end{table*}

\subsection{Prototype-Centric Sampling}
In order to better adapt to unlabeled data, the TTA method optimizes the model by reducing self-supervised loss, which makes the quality of the samples fed into the loss function crucial. 
As a result, we aim to improve the TTA strategy by proposing a new sampling method.
Considering that frames with high similarity to the prototype $\mathbf{y}$ are more likely to belong to action \( \hat{c} \), 
sampling \( K \) visual features from moments with the highest similarity to the prototype, rather than selecting them randomly, can provide more informative visual cues. 
This approach, referred to as \textit{Prototype-Centric Sampling (PCS)}, enables more effective alignment of visual and textual features from ViL models within the semantic space. The performance of different sampling strategies is analyzed in Section \ref{sec:samplingway}
\section{Experiment}
\label{sec:experiment}
\subsection{Dataset and Metrics}

We evaluate our method on two popular TAD benchmarks: THUMOS14 \cite{idrees2017thumos} and ActivityNet-1.3 \cite{caba2015activitynet}. The THUMOS14 covers 20 action categories with 200 training videos and 213 testing videos. The ActivityNet-1.3 dataset, collected from YouTube, comprises 200 human action categories and contains a total of 19,994 videos. These videos are divided into three sets: 10,024 for training, 4,926 for validation, and 5,044 for testing. 

For ZSTAD evaluation, we follow the standard setting \cite{ju2022prompting,nag2022zero,liberatori2024test} and split the data according to 75\%-25\%  and 50\%-50\%. To ensure statistical significance, we randomly sample 10 times under each split strategy and take the average as the final results.
Following the standard protocols, the mean average precision (mAP) at different temporal intersection over union (tIoU) thresholds is used as main evaluation metrics. Given a tIoU threshold $\xi$, the mean of average precision across all action categories denoted as mAP$_\xi$ is computed. Unless specifically noted, mAP refers to an average of mAP values across different tIoUs.

\subsection{Implementation Details}

For a fair comparison, we adopt a multimodal large-scale model CoCa (ViT-L/14) \cite{yu2022coca} as backbone following \cite{liberatori2024test} for visual embedding and textual encoding. When extracting features, the RGB frames across all videos maintain the original rate, and a textual prompt with the template ``a video of action \textit{CLS}" is used for any class name of interest. The adaptation is conducted with the popular Adam optimizer at a learning rate of 1e-5 and a weight decay of 1e-4. The hyperparameters $T$ and $K$ are set to 60/25 and 4/20 for THUMOS14 dataset and ActivityNet-1.3 dataset, respectively. The value for $\beta$, $\gamma_1$, and $\gamma_2$ are 1, 0.2, and 1 for both datasets. Additionally, we replace the fixed threshold $\alpha$ with the average similarity score across the entire video. 
All of our experiments are conducted on a single NVIDIA RTX 4090 GPU with 24 GB of memory.

\subsection{Experimental Results} 
\noindent\textbf{Performance Comparison:}
We categorize existing literature on ZSTAD into two groups based on learning strategies: (1) supervised learning, which uses standard training and testing with fully annotated data, as in Eff-Prompt \cite{ju2022prompting}, STALE \cite{nag2022zero}, and DeTAL \cite{li2024detal}; and (2) unsupervised learning, which operates without any annotations, exemplified by T3AL \cite{liberatori2024test}, along with our methods, FreeZAD and AdaZAD.
Considering the limited number of methods that abstain from training on labeled data, along with the intuitive performance gains contributed by our models, we establish a naive baseline model, a straightforward architecture that detects activities by simply calculating the similarity between visual and textual features extracted by their respective backbones.

Table \ref{table1} illustrates quantitative comparisons between our methods, the naive baseline, and state-of-the-art methods on open-set scenarios of TAD.
All results indicate that existing large-scale model applications are insufficient for the ZSTAD task. Under the zero-shot setting, unsupervised methods using video-level pseudo-labeling perform worse compared to fully-supervised methods trained with precise annotations, and this phenomenon is also evident in close-set scenarios \cite{zhang2024hr}. 
However, our adaptation-based method considerably narrows this gap. For example, AdaZAD outperforms Eff-Prompt \cite{ju2022prompting} by 0.4\% average mAP on ActivityNet-1.3 under 50\%-50\% split, which learns continuous prompt vectors to reduce task discrepancies. 

Due to its design tailored for the absence of explicit temporal modeling and annotated information, our FreeZAD achieves an average mAP improvement of 2.8\% over the naive baseline on the THUMOS14 dataset under a 75\%-25\% split. 
When coupled with TTA, this improvement increases to 6.8\%.
Similar trends are observed across various datasets and splits. 
Compared to other unsupervised learning method, our FreeZAD achieves comparable performance with adaptation-based T3AL \cite{liberatori2024test} by +0.8\%/-0.5\% mAP on THUMOS 14 and surpasses it by 2.9\%/4.4\% mAP on ActivityNet-1.3 under 75\%-25\%/50\%-50\% splits, respectively, without any fine-tuning or adaptation. 
These results confirm that our model effectively harnesses the generalization capabilities of ViL models for ZSTAD.
When equipped with our designed TTA, AdaZAD significantly outperforms T3AL, with improvements of 4.8\%/3.7\% on THUMOS14 and 3.9\%/5.7\% on ActivityNet-1.3. This outcome underscores the adaptability of FreeZAD and validates the effectiveness of our TTA strategy.

\begin{table}[t]
\setlength{\tabcolsep}{3.5mm}
\begin{tabular}{c|ccc|c}
\toprule[1pt]
\midrule
Method & Param. & Mem. & FPS   & mAP  \\ \midrule
T3AL\cite{liberatori2024test}   & 1.2M   & 8.6G & 10.1  & 9.2  \\
FreeZAD    & 0      & 3.3G & 128.9 & 10.0 \\
AdaZAD & 1.2M   & 3.7G & 28.4  & 14.0 \\ \midrule
\bottomrule[1pt]
\end{tabular}
\caption{\textbf{Runtime comparison.} Param., Mem., and FPS denote the number of learnable parameters, GPU memory usage and frames per second during inference, respectively.}
\label{table4}
\end{table}

\noindent\textbf{Runtime Comparison:} 
Table \ref{table4} presents the runtime comparison of our training-free and adaptation-based methods alongside the state-of-the-art unsupervised competitor \cite{liberatori2024test} on the THUMOS14 dataset with a 75\%-25\% split. FreeZAD achieves 0.8 mAP higher than T3AL while using only around \textit{1/13} of the runtime. This remarkable efficiency is attributed to its simple yet effective design, eliminating the computational burden of additional fine-tuning or adaptation. With TTA enabled, AdaZAD outperforms T3AL by a substantial margin (4.8\%) while maintaining approximately 2.8 times the speed of T3AL.
\subsection{Ablation Studies and Model Analyses}

In this section, a quantitative analysis is conducted to evaluate the effectiveness of various model components, explore the importance of adaptation steps, validate the potential of AdaZAD, and  and analyze error types. Unless otherwise specified, all experiments are conducted on the THUMOS14 dataset under 75\%-25\% split.

\begin{table}[t]
\setlength{\tabcolsep}{1mm}
\begin{tabular}{cc|ccccc|c}
\toprule[1pt]
\midrule
Score                       & \begin{tabular}[c]{@{}c@{}}Actionness\\ Calibration\end{tabular} & 0.3  & 0.4  & 0.5  & 0.6 & 0.7 & mAP \\ \midrule
\multirow{2}{*}{Similarity} & \ding{55}                                                               & 20.5 & 13.4 & 8.4  & 4.7 & 2.4 & 9.9    \\
                            & \ding{51}                                                               & 22.1 & 14.6 & 9.4  & 5.4 & 2.8 & 10.9     \\ \midrule
\multirow{2}{*}{OIC}        & \ding{55}                                                               & 23.3 & 15.4 & 9.7 & 4.9 & 2.5 & 11.2    \\
                          & \ding{51}                                                               & 25.0 & 16.7 & 10.6 & 5.9 & 3.1 & 12.3    \\ \midrule
\multirow{2}{*}{LogOIC}    & \ding{55}                                                               & 26.4 & 17.7 & 11.4 & 6.1 & 3.1 & 12.9    \\
                            & \ding{51}                                                               & 27.5 & 18.7 & 12.3 & 7.3 & 4.0 & 14.0    \\ \midrule
\bottomrule[1pt]
\end{tabular}
\caption{\textbf{Ablation studies on the confidence score and actionness calibration,} in which our LogOIC coupled with actionness calibration outperforms all other cases.}
\label{table_score}
\end{table}

\begin{table}[t]
\setlength{\tabcolsep}{1.4mm}
\begin{tabular}{cc|ccccc|c}
\toprule[1pt]
\midrule
\begin{tabular}[c]{@{}c@{}}Positive\\ Sample\end{tabular} & \begin{tabular}[c]{@{}c@{}}Negative\\ Sample\end{tabular} & 0.3  & 0.4  & 0.5  & 0.6 & 0.7 & mAP \\ \midrule
\ding{55}                                                         & \ding{55}                                                         & 25.3 & 17.8 & 10.7 & 6.4 & 3.4 & 12.6    \\
\ding{51}                                                         & \ding{55}                                                         & 27.5 & 18.7 & 12.3 & 7.3 & 4.0 & 14.0    \\
\ding{55}                                                         & \ding{51}                                                         & 23.5    & 15.0    & 9.3    & 5.4   & 2.9   & 11.2       \\ \midrule
\bottomrule[1pt]
\end{tabular}
\caption{\textbf{Ablation studies on the sampling way.} Positive/negative samples marked with \ding{51} are derived from frames closest/furthest to the prototype, while those marked with \ding{55} are randomly chosen based on similarity scores.}
\label{table_sampling}
\end{table}

\noindent\textbf{Impact of Confidence Score:} We assess the impact of different scores in our AdaZAD model and evaluate the effectiveness of actionness calibration. The results are summarized in Table \ref{table_score}. The similarity score is obtained by calculating the cosine similarity between the activity textual features and the region-level visual features derived from the visual representations of each temporal segment. 
Regardless of whether actionness calibration is applied, LogOIC outperforms both similarity and OIC by over 3\% and 1.7\% in average mAP, thanks to its well-designed structure that improves boundary sensitivity.
Additionally, examining the statistics of different confidence types reveals that the actionness calibration technique refines confidence scores effectively and improves the model’s predictive accuracy by incorporating frequency energy.

\noindent\textbf{Impact of Sampling Strategies:}
\label{sec:samplingway}
We experiment with various positive and negative sampling strategies, demonstrating that sample selection significantly influences model adaptation, as shown in Table \ref{table_sampling}. Positive/negative samples marked with \ding{51} are derived from frames closest/furthest to the textual feature of pseudo-label in semantic space, while those marked with \ding{55} are randomly chosen based on similarity scores. Selecting $K$ positive samples randomly near the semantic closest points can enhance the generalization performance of the model with 1.4\% mAP. However, this sampling way is not suitable for the negatives. This is because samples near the furthest frame from the prototype are easy negatives, which can hardly enhance the capability of the model to distinguish challenging samples.
Overall, our PCS approach for selecting key positive sample features enhances the model’s localization accuracy during adaptation.

\begin{figure}[t]
    \centering
    \begin{subfigure}[b]{0.22\textwidth}
        \centering
        \includegraphics[width=\textwidth]{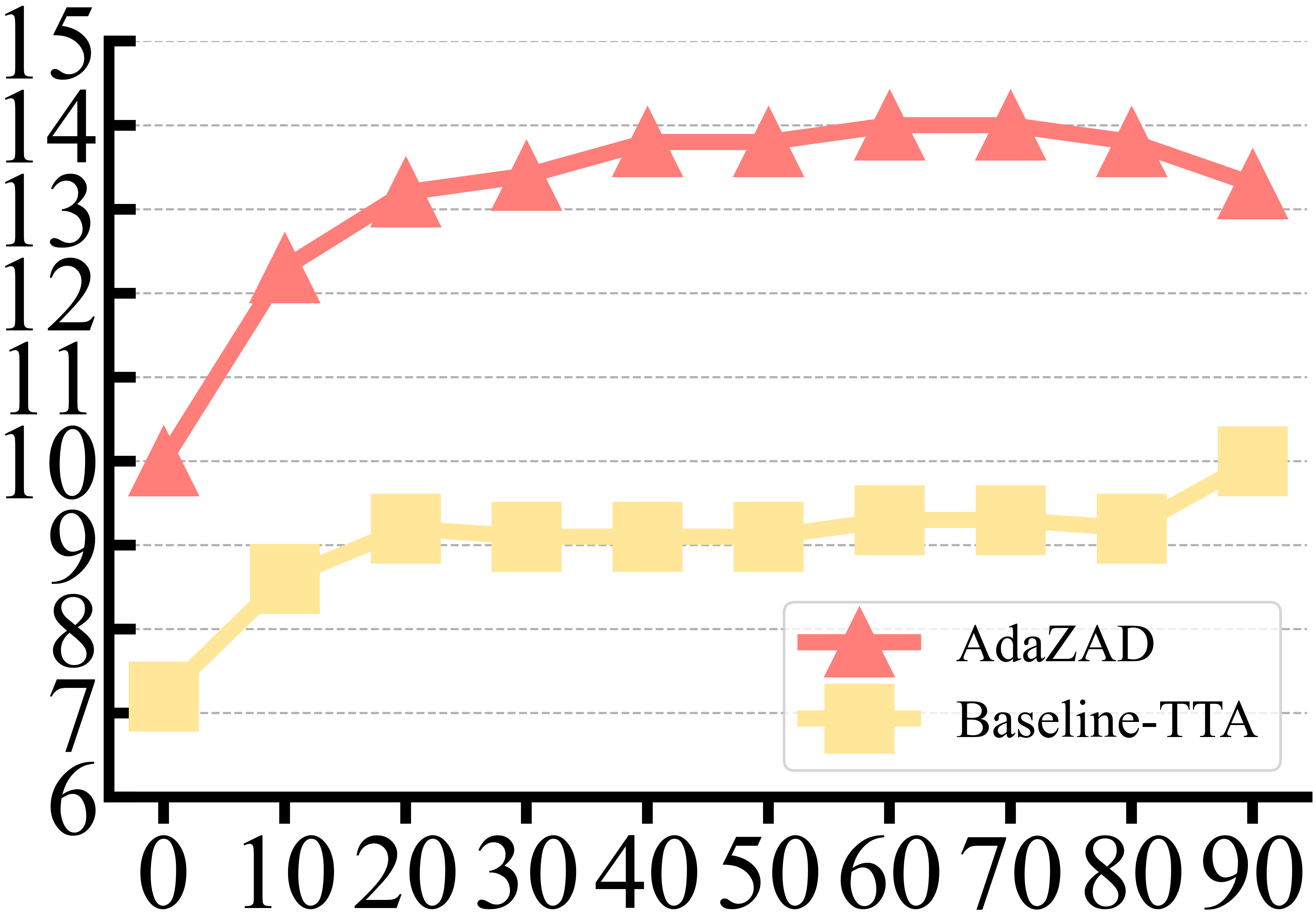}
        \caption{Different adaptation steps $M$}
        \label{fig4a}
    \end{subfigure}
    \hspace{0.02\textwidth}
    \begin{subfigure}[b]{0.22\textwidth}
        \centering
        \includegraphics[width=\textwidth]{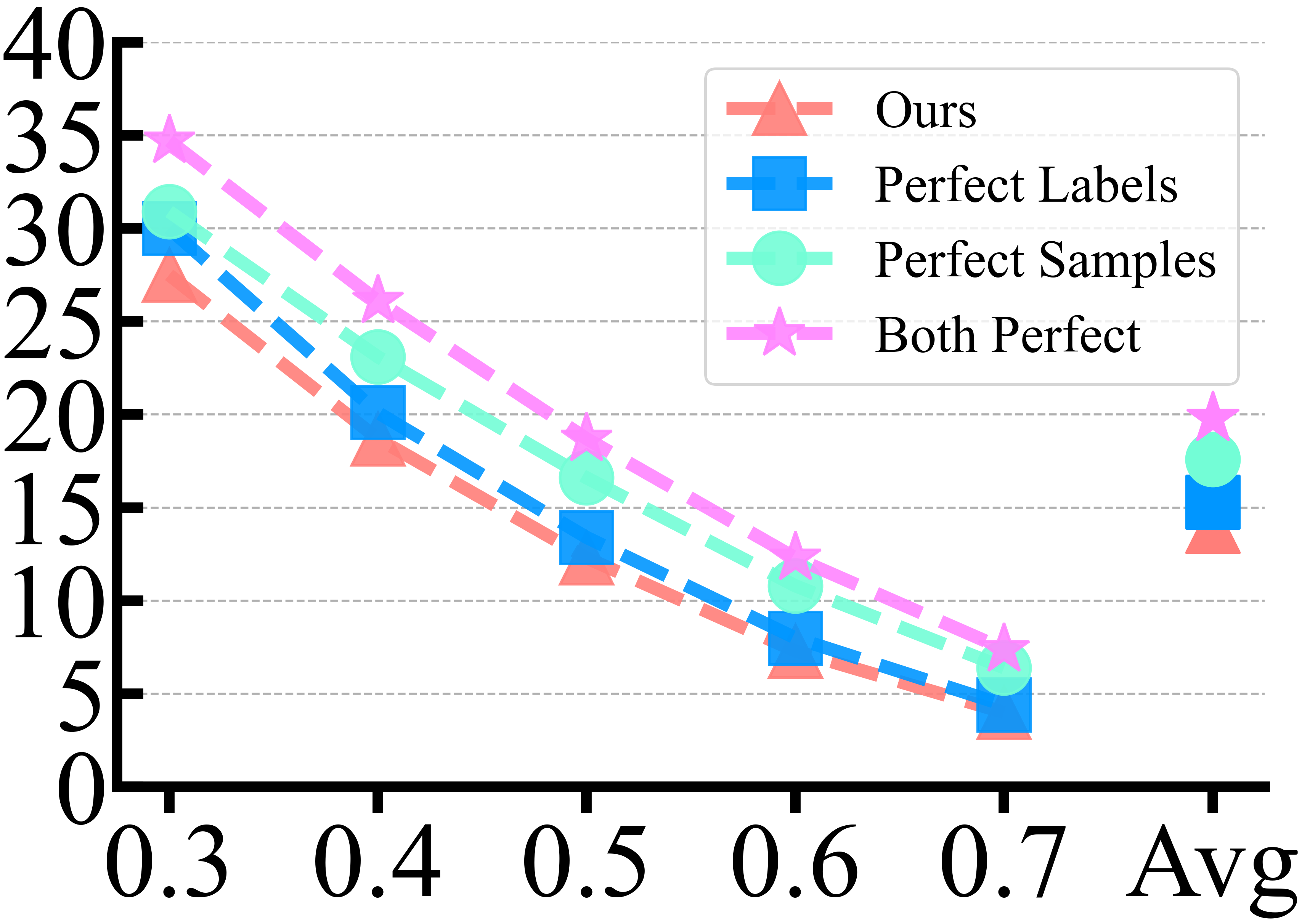}
        \caption{Different tIoU thresholds}
        \label{fig4b}
    \end{subfigure}
    \caption{The left plot shows the average mAP across various adaptation steps $M$ , while the right plot presents the average mAP for different tIoU thresholds under three relaxed unsupervised constraints: perfect class, perfect samples, and a combination of both.}
    \label{fig_ablation}
\end{figure}

\noindent\textbf{Impact of Adaptation Step:} Adapting the model to unlabeled data is crucial for the ZSTAD task, where the hyperparameter $M$ controls the number of adaptation steps, playing a significant role in determining the model’s performance.
We keep all settings of the Baseline with TTA and our AdaZAD fixed except for adaptation steps \( M \), which range from 0 to 90 with a stride of 10. 
As observed from the orange curve in Figure \ref{fig_ablation} (a), the mAP of AdaZAD increases by 2.3\% when \( M = 10 \) compared to $M=0$. As \( M \) increases, the performance of the model increases gradually until \( M = 60 \), accompanied by a diminishing marginal effect. However, when \( M \geq 80 \), performance declines as the model’s continuous updates gradually amplify the impact of noise from the single video sequence.
To balance effectiveness and efficiency, we set \( M = 60 \) for the comparison experiments with peer work. 
Another interesting finding is that when $M=0$, AdaZAD (\ie, FreeZAD) achieves performance comparable to Baseline-TTA at $M=90$, demonstrating the superiority of our training-free strategy.

\noindent\textbf{Oracle Analysis:} To validate the potential of our method, we employ different types of oracle information, including perfect pseudo-labels, perfect samples, and combination of both perfect settings to replace the corresponding components in our model for further evaluation.
As shown in Figure~\ref{fig_ablation} (b), perfect labels refer to pseudo-labels that are completely correct, and perfect samples are positive/negative samples derived from the real foreground/background of the video. We find that, under these two configurations, the average mAP of our model improves by 1.3\% and 3.6\%, respectively. When combining both types of oracle information, the mAP increases by up to 5.8\%. All these results demonstrate the significant potential of our model.

\begin{figure}[t]
\centering
\includegraphics[width=1\columnwidth]{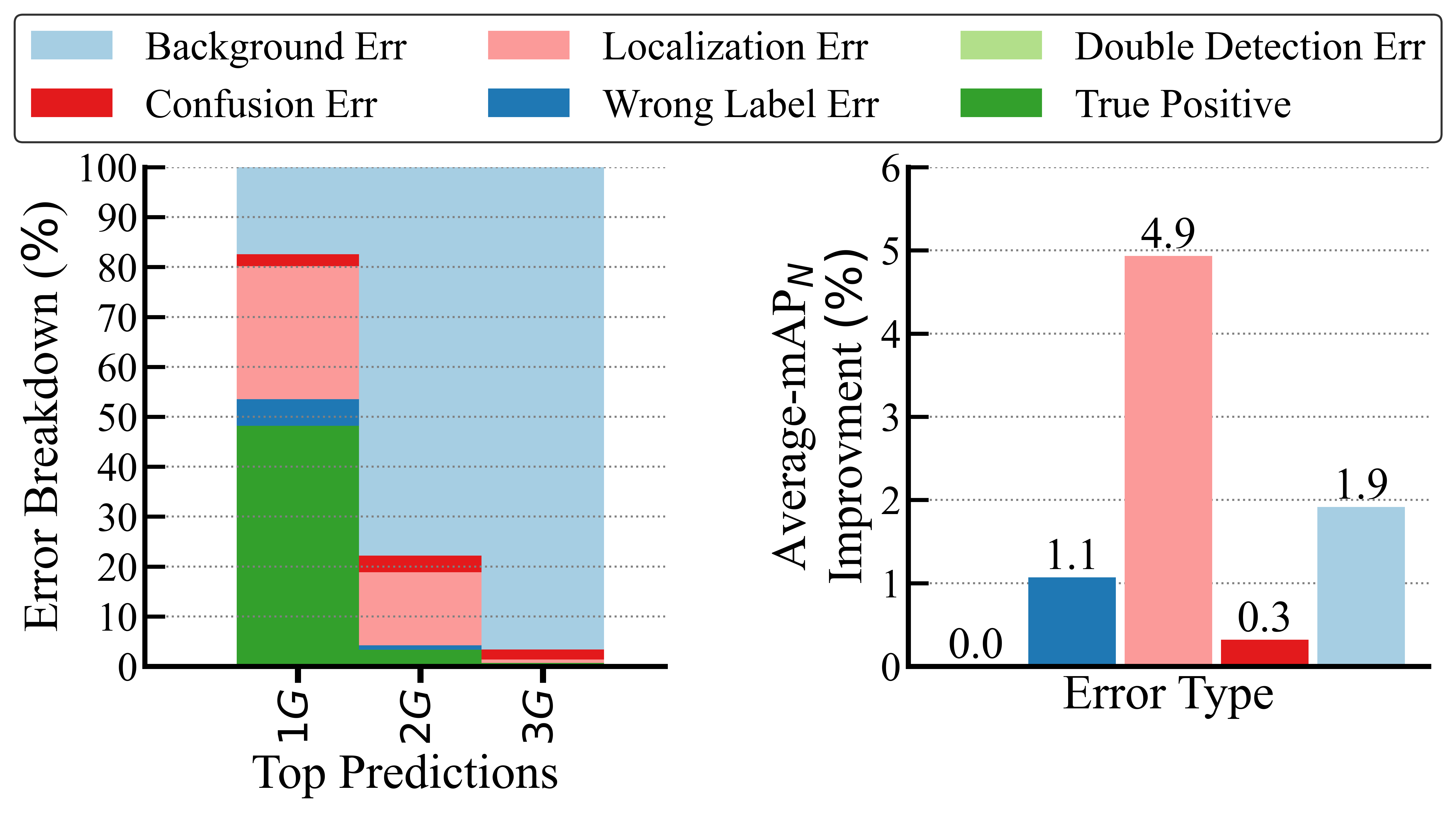} 
\caption{\textbf{Error analysis of our FreeZAD.} The left is false positive profile and the right is removing error impact, where $G$ denotes the number of ground truths. Only the Top-1G to Top-3G predictions are provided due to the sparsity of the predictions.}
\label{fig_fpa}
\end{figure}

\noindent\textbf{Error Analysis:}
To assess the limitations of our method, we conduct a false positive analysis \cite{alwassel2018diagnosing} at tIOU=0.5 on the THUMOS14 dataset. Given that our method generates sparse predictions, we report results for Top-1G to Top-3G predictions, as shown in Figure \ref{fig_fpa}. On Top-1G predictions, our approach achieves a high true positive rate but shows some weaknesses in localization accuracy.
Since our method produces non-overlapping segments, it effectively avoids double-detection errors across predictions. 
However, as an anchor-free method, it is more susceptible to background errors due to the lack of anchor-based guidance, which results in a higher likelihood of unmatched predictions relative to the ground truth. 

\section{Conclusion}
By leveraging ViL models, we propose FreeZAD, a training-free approach for zero-shot temporal action detection. 
Extensive evaluations on two popular benchmark datasets, THUMOS14 and ActivityNet-1.3, demonstrate that our training-free method achieves performance comparable to the state-of-the-art unsupervised method T3AL, while requiring only about \textit{1/13} of the inference time. The enhanced model with a designed test-time adaptation strategy further outperforms the performance and narrows the gap with fully supervised methods. 
Extensive ablation studies indicate that our Logarithmic Decay Weighted Outer-Inner-Contrastive Score and Actionness Calibration can reliably rank predictions and significantly outperform similarity-based approaches. Additionally, the proposed Prototype-Centric Sampling enables the model to capture more valuable information, enhancing alignment between visual and textual features. Furthermore, we conduct an oracle analysis by relaxing certain unsupervised constraints, demonstrating the potential of our model and indicating future directions of improvement.
{
    \small
    \bibliographystyle{ieeenat_fullname}
    \bibliography{main}
}

\end{document}